\pdfoutput=1

\documentclass[11pt]{article}

\usepackage[]{EACL2023}

\usepackage{times}
\usepackage{latexsym}
\usepackage{xcolor}

\usepackage{amsmath}
\usepackage{graphicx}
\usepackage{verbatim}
\usepackage[noend]{algpseudocode}
\usepackage{algorithmicx,algorithm}

\usepackage[T1]{fontenc}

\usepackage[utf8]{inputenc}

\usepackage{microtype}

\usepackage{inconsolata}

\title{Better Pre-Training by Reducing Representation Confusion}

\begin{document}
\author{Haojie Zhang\textsuperscript{1, 2}\footnotemark[1], 
Mingfei Liang\textsuperscript{1}\footnotemark[1], 
Ruobing Xie\textsuperscript{1}\footnotemark[1], 
Zhenlong Sun\textsuperscript{1}, 
Bo Zhang\textsuperscript{1}, Leyu Lin\textsuperscript{1} \\  \textsuperscript{1}WeChat Search Application Department, Tencent, China\\
\textsuperscript{2}Peking University, China\\ 
 \textsuperscript{1} \{coldhjzhang, aesopliang, ruobingxie, richardsun, nevinzhang, goshawklin\}@tencent.com \\
 \textsuperscript{2}zhanghaojie@stu.pku.edu.cn 
 }

\maketitle
\renewcommand{\thefootnote}{\fnsymbol{footnote}}
 \footnotetext{\footnotemark[1]Equal contribution.}
 \renewcommand{\thefootnote}{\arabic{footnote}}
\begin{abstract}
In this work, we revisit the Transformer-based pre-trained language models and identify two different types of information confusion in position encoding and model representations, respectively. 
Firstly, we show that in the relative position encoding, the joint modeling about relative distances and directions brings confusion between two heterogeneous information. It may make the model unable to capture the associative semantics of the same distance and the opposite directions, which in turn affects the performance of downstream tasks. Secondly, we notice the BERT with Mask Language Modeling (MLM) pre-training objective outputs similar token representations (last hidden states of different tokens) and head representations (attention weights\footnote{"attention weights" mainly refer to the dot product between Key and Query in the self-attention module.}of different heads), which may make the diversity of information expressed by different tokens and heads limited.
Motivated by the above investigation, we propose two novel techniques to improve pre-trained language models: Decoupled Directional Relative Position (DDRP) encoding and MTH\footnote{MTH is the abbreviation of our proposed MLM with Token Cosine Differentiation (TCD) and Head Cosine Differentiation (HCD) pre-training task. TCD and HCD are described in detail in sec.	\ref{sec.introduction}(2) and sec.\ref{sec.CD}.} pre-training objective. 
DDRP decouples the relative distance features and the directional features in classical relative position encoding. 
MTH applies two novel auxiliary regularizers besides MLM to enlarge the dissimilarities between (a) last hidden states of different tokens, and (b) attention weights of different heads. 
These designs allow the model to capture different categories of information more clearly, as a way to alleviate information confusion in representation learning for better optimization.
Extensive experiments and ablation studies on GLUE benchmark demonstrate the effectiveness of our proposed methods. 
\end{abstract}

\section{Introduction}
\label{sec.introduction}
The paradigm of pre-training on large-scale corpus and fine-tuning on specific task datasets has swept the entire field of Natural Language Processing (NLP). BERT \cite{devlin2018bert} is the most prominent pre-trained language model, which stacks the encoder blocks of Transformer \citep{vaswani2017attention} and adopts MLM and Next Sentence Prediction (NSP) pre-training tasks, achieving the SOTA results in 2018. After that, a large number of Pre-trained Language Models (PLMs) \cite{liu2019roberta,lan2019albert,raffel2019exploring,clark2020electra,he2020deberta} that optimize the Transformer structure and pre-training objectives have emerged, which further improves the performance of the pre-trained language models on multiple downstream tasks.
In this work, we identify two different types of information confusion in language pre-training, and explore two conceptually simple and empirically powerful techniques against them as follows:

(1) \textbf{Decoupled Directional Relative Position (DDRP) Encoding}. 
It is well known that relative position encoding is competitive and has been widely used in real PLMs \cite{shaw2018self,yang2019xlnet,wei2019nezha,raffel2019exploring,su2021roformer,he2020deberta,ke2020rethinking}. Despite its great performance, we still notice relative position encoding methods utilizes completely separate parametric vectors to encode different relative position information, which indicates that every single parametric vector needs to learn both distance and directional features.
We consider this paradigm of utilizing a single parametric vector to represent both relative distance and direction as a kind of information confusion, and question its rationality.
Since relative distance features and the directional features are apparently heterogeneous information that reflects different aspects of positional information, we argue that existing methods may impose difficult in establishing connections explicitly between parametric vectors of the same distances and the opposite directions, which in turn result in serious information losses in position encoding. 
Inspired by this, we propose a novel Decoupled Directional Relative Position (DDRP) encoding. In detail, DDRP decomposes the classical relative position embedding \cite{shaw2018self} into two embeddings, one storing the relative distance features and the other storing the directional features, and then multiply the two together explicitly to derive the final decoupled relative position embedding, allowing originally confused distance and directional information to be as distinguishable as possible.

(2) \textbf{Model Representation Differentiations}. 
We  analyze that there is non-negligible confusion in the representation of pre-trained BERT,
as evidenced by the high consistency in last hidden states across different tokens and attention weights across different heads, respectively. 
Similar last hidden states will introduce the \emph{anisotropic problem} \cite{mimno-thompson-2017-strange}, which will bound the token vectors to a narrow representation space and thus make it more difficult for the model to capture deep semantics. 
Considering attention weights contain rich linguistic knowledge \cite{clark2019does,jawahar2019does}, we argue that high consistency in attention weights also constrains the ability of the model to capture multi-aspect information. 
To alleviate the representation confusion between different tokens and heads caused by high information overlap, we propose two novel pre-training approaches to stimulate the potential of the pre-trained model to learn rich linguistic knowledge: Token Cosine Differentiation (TCD) objective and Head Cosine Differentiation (HCD) objective. 
Specifically, TCD attempts to broaden the dissimilarity between tokens by minimizing the cosine similarities between different last hidden states. 
In contrast, HCD attempts to broaden the dissimilarity between heads by minimizing the cosine similarities between different attention weights. 
We apply  TCD and HCD as two auxiliary regularizers in MLM pre-training, which in turn guides the model to produce more discriminative token representations and head representations. 
Formally, we define our enhanced pre-training task as \textbf{M}LM with \textbf{T}CD and \textbf{H}CD (\textbf{MTH}).

Extensive experiments on the GLUE benchmark show that DDRP achieves better results than classical relative position encoding \cite{shaw2018self} on almost all tasks without introducing the additional computational overhead and consistently outperforms prior competitive relative position encoding models \cite{he2020deberta,ke2020rethinking}. Moreover, our proposed MTH outperforms MLM  by a 0.96 average GLUE score and achieves nearly 2x pre-training speedup on BERT\(_{BASE}\). Both DDRP and MTH are straightforward, effective, and easy to deploy, which can be easily combined with existing pre-training objectives and various model structures.
Our contributions are summarized as follows: 
\begin{itemize}
    \item We propose a novel relative position encoding named DDRP, which decouples the relative distance and directional features, giving the model a stronger prior knowledge, fewer parameters, and better results compared to conventional coupled position encodings.
    \item We analyze the trend of self-similarity of last hidden states and attention weights during pre-training, and propose two novel Token Cosine Differentiation and Head Cosine Differentiation objectives, motivating pre-trained Transformer to better capture semantics in PLMs.
    \item  We experimentally verified by our proposed techniques (DDRP and MTH) that decomposing heterogeneous information and extending representation diversity can significantly improve pre-trained language models. We also analyze the characteristics of DDRP and MTH in detail.
\end{itemize}

\section{Related Work}
In recent years, pre-trained language models have made significant breakthroughs in the field of NLP. BERT \citep{devlin2018bert}, which proposes MLM and NSP pre-training objectives, is pre-trained on large-scale unlabeled corpus and has learned bidirectional representations efficiently. After that, many different pre-trained models are produced, which further improve the effectiveness of the pre-trained models. RoBERTa \citep{liu2019roberta} proposes to remove the NSP task and verifies through experiments that more training steps and larger batches can effectively improve the performance of the downstream tasks. ALBERT \citep{lan2019albert} proposes a Cross-Layer Parameter Sharing technique to lower memory consumption. XL-Net \citep{yang2019xlnet} proposes Permutation Language Modeling to capture the dependencies among predicted tokens. ELECTRA \citep{clark2020electra} adopts Replaced Token Detection (RTD) objective, which considers the loss of all tokens instead of a subset.  TUPE \citep{ke2020rethinking} performers Query-Key dot product with different parameter projections for contextual information and positional information separately and then added them up, they also add relative position biases like T5 \cite{raffel2019exploring} on different heads to form the final correlation matrix. DEBERTA \citep{he2020deberta} separately encodes the context and position information of each token and uses the textual and positional disentangled matrices of the words to calculate the correlation matrix.

\section{Method}
In this section, we analyze in turn two different types of information confusion that exist in the real PLMs: (i) The paradigm of utilizing a single parametric vector of relative position embedding to represent both relative distance and direction. (ii) The high similarity and overlap in model representations. Based on above two investigations, we propose two techniques, \textbf{D}ecoupled \textbf{D}irectional \textbf{R}elative \textbf{P}osition (\textbf{DDRP}) Encoding and \textbf{M}LM with \textbf{T}CD and \textbf{H}CD (\textbf{MTH}), respectively, to help the PLMs alleviate information confusion and enhance representation clarity and diversity.

\subsection{Decoupled Directional Relative Position (DDRP) Encoding}
\label{sec.DDRP}
We first start to introduce DDRP by formulating multi-head attention module of BERT and BERT-R \cite{shaw2018self}. Specifically, BERT formulates multi-head attention for a specific head as follows:

\begin{footnotesize}\begin{gather}
Q=HW^Q,K=HW^K,V=HW^V,  \tag{1}\\
A=\frac{Q{K}^T}{\sqrt d}, \tag{2}\\
Z=softmax\left(A\right)V,\tag{3}\end{gather}\end{footnotesize} where \(H \in R^{S\times D}\) represents the input hidden states; \(W^Q\), \(W^K\), \(W^V\) \(\in R^{D\times d}\) represent the projection matrix of Query, Key, and Value respectively; \(A \in R^{S\times S} \) represents attention weight; \(Z \in R^{S\times d}\) represents the single-head output hidden states of self-attention module; \(S\) represents input sequence length; \(D\) represents the dimension of input hidden states; \(d\) represents the dimension of single-head hidden states.
Unlike BERT, which adds the absolute position embedding to the word embedding as the final input of the model, BERT-R first applies relative position encoding. It 
adds relative position embedding into \(K\) in the self-attention module of each layer to make a more interactive influence. Its formulations are as follows:

\begin{footnotesize}\begin{gather}A_{i,j}=\frac{Q_{i}\left(K_{j}+K_{\sigma\left(i,j\right)}^r\right)^T}{\sqrt d} \tag{4}, \\
\sigma\left(i,j\right)=clip\left(i-j\right)+r_s, \tag{5} 
\end{gather}\end{footnotesize}where  \(Q_{i}\) represents Query vector at the i-th position; \(K_{j}\) represents Key vector at the j-th position; \(r_s\) represents maximum relative position distance; \(\sigma\left(i,j\right)\) represents the index of relative position embedding \(K^r \in R^{2r_s \times d}\); relative position embedding for \(K\) are shared at all different heads. Note that \citet{shaw2018self} has experimentally demonstrated that adding relative position embedding to the interaction between \(A\) and \(V\) gives no further improvement in effectiveness, so the relative position embedding in \(V\) space is eliminated in all our experiments to reduce the computational overhead.

\begin{figure}[htp]
    \centering
    \includegraphics[width=7.5cm]{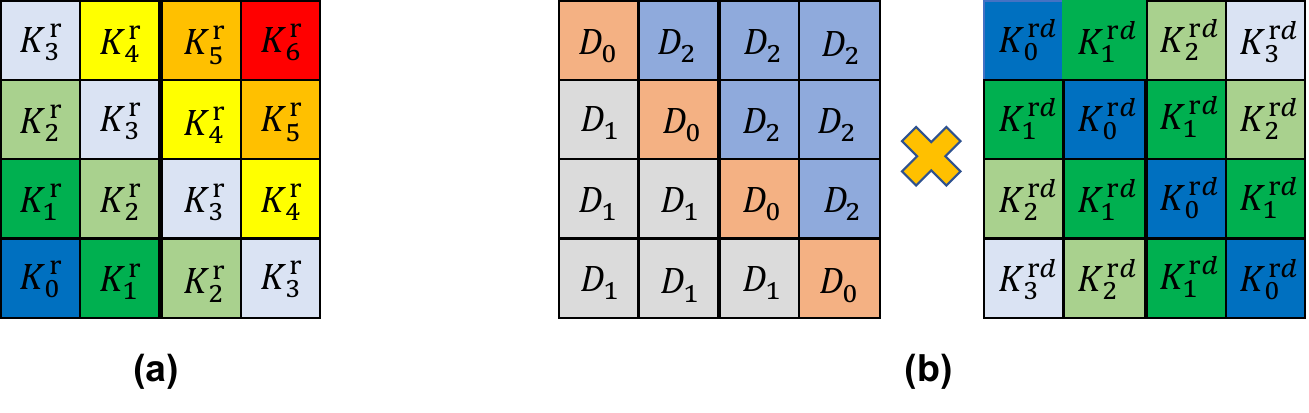}
    \caption{Fig.1(a) represents the classical relative position matrice; Fig.1(b) represents the decoupled relative position matrices we proposed. Note that the  parametric vectors of the same color have the same values.}
    \label{fig:one}
\end{figure}

Compared with BERT, BERT-R models the correlation between words and positions more explicitly, and thus further expands the expression diversity between words. However, we notice that in BERT-R, the vectors from the same distance on both left and right sides are encoded in isolation (as shown in Figure~\ref{fig:one}(a)), which indicates that every single parametric vector from \(K^r\) is forced to maintain distance and direction, two different types of information. Since it is confirmed that directional information is crucial in language modeling \cite{vu2016bi,fuller2002arrow,shen2018disan}, we argue that such an approach causes unnecessary information confusion and faces several constraints: (i) Mixing relative distance and directional information for modeling makes information originally in different spaces entangled, which in turn makes the learning of parametric vectors more difficult. (ii) Dot products between word vectors and directionally confused positional vectors bring unnecessary randomness in deep bidirectional representation models.

To alleviate the confusion of distance and direction that exists in BERT-R and allow the model to perceive distances and directions more clearly, we propose a novel Decoupled Directional Relative Position (DDRP) encoding. Specifically, DDRP decouples the relative distance and directional information and maintains them with two different embeddings. Its formula is as follows:

\begin{footnotesize}\begin{gather}
A_{i, j}=\frac{Q_{i}\left(K_{j}+K_{\delta(i, j)}^{d}\right)^T}{\sqrt{d}} \tag{9},\\
K^{d}_{\delta\left(i,j\right)}=D_{\rho\left(i,j\right)} \odot K_{\delta\left(i,j\right)}^{rd}, \tag{10} \\
\rho\left(i,j\right) =\left\{\begin{matrix} 
  1,\ \text{\(if \ i-j<0\)} \\ 
  0,\ \text{\(if  \ i-j=0\)} \\ 
  2,\ \text{\(if  \ i-j>0\)} 
\end{matrix}\right. ,\tag{11}\\
\delta(i, j)=a b s(c l i p(i-j)),\tag{12}\label{Equation13}
\end{gather}\end{footnotesize} where \(\rho\left(i,j\right)\) represents the index of directional embedding \(D \in R^{3\times d}\); \(\delta\left(i,j\right)\) represents the index of  relative distance embedding \(K^{rd}\in R^{r_s\times d}\); \(K^{d}\) represents the relative position matrice. Note that in terms of implementation details, the only difference between DDRP and BERT-R is that DDRP decouples \(K^r\) in BERT-R into the element-wise multiplication of \(D\) and \(K^{rd}\). We also provide a specific comparison example in Figure~\ref{fig:one}.

Compared to previous relative position encodings, we summarize the advantage of DDRP as follows: (i) DDRP explicitly extracts the commonalities (relative distances) and differences (directions) in the positional information, leading the model to produce attention that better match the real semantic distributions, which reduces the difficulty of model learning and unlocks the potential of the model. (ii) DDRP compresses the total number of parametric vectors from \({2r}_s\) to \({r}_s + 3\).
\subsection{Model Representation Differentiations}
\label{sec.CD}
\textbf{Token representations.}
Isotropic distributions have been proved theoretically to be beneficial to token representations, which ensures that the different token vectors are directional uniform, thus maximizing the diversity of token representations. \cite{mimno-thompson-2017-strange}. In practice, \citet{mu2017all} have also empirically confirmed the effectiveness of isotropic distributions on static token representations, such as WORD2VEC \cite{mikolov2013efficient} and GLOVE \cite{pennington2014glove}. Inspired by the above studies, we also wonder whether contextualized token representations (e.g., last hidden states of BERT) are isotropic. Following \citet{mimno-thompson-2017-strange}, we utilize the cosine similarity to evaluate the degree of isotropy in token representations. The higher similarity, the smaller isotropy; the lower similarity, the greater isotropy. For an input sequence \(S=[x_1,\ldots ,x_n]\), we formulate the last hidden states' average self-similarity as follows:

\begin{footnotesize}\begin{gather}
f(S)=\frac{2}{n(n-1)} \sum_{i=1}^{n-1} \sum_{j=i+1}^{n} \cos \left(h_{i}, h_{j}\right), \tag{15}\end{gather}\end{footnotesize}where \(h_i\) and \(h_j\) are the last hidden states of \(x_i\) and \(x_j\); \(cos\) represents cosine similarity.

\noindent{\textbf{Head representations.}}
Multi-head attention, is aimed at capturing information in different heterogeneous subspaces and has been experimentally verified different heads correspond
well to different linguistic notions \cite{clark2019does}. However, some studies point out that some heads contribute almost nothing to downstream tasks \cite{kovaleva2019revealing,michel2019sixteen,voita-etal-2019-analyzing,correia-etal-2019-adaptively}. We are surprised by this, and speculate that the above problem may be caused by the heavy overlap of information that some heads are concerned about. 
To verify our point, we utilize cosine similarity to evaluate the degree of overlap in head representations, following token representations. The higher similarity, the higher overlap; the lower similarity, the lower overlap.
For multiple heads \(H=[H_1^1,\ldots,H_m^1,\ldots,H_1^L,\ldots,H_m^L]\), we formulate the attention weights' average self-similarity as follows:

\begin{footnotesize}
\begin{gather}
f(H)=\frac{2}{Lm(m-1)} \sum_{l=1}^L\sum_{i=1}^{m-1} \sum_{j=i+1}^{m} \cos \left(a_{i}^l, a_{j}^l\right), \tag{16}
\end{gather}\end{footnotesize}where \(L\) represents the number of Transformer layers; \(a_{i}^l\) and \(a_{j}^l\) are the attention weights of the i-th head and j-th head of the l-th layer. 

\noindent{\textbf{Analysis on the similarities between different tokens and heads.} }
With curiosity about the similarity of token representations and head representations, we analyze the self-similarity trends of tokens and heads during the original MLM BERT pre-training.  
\begin{figure}[htpb]
    \centering
    \includegraphics[width=7cm]{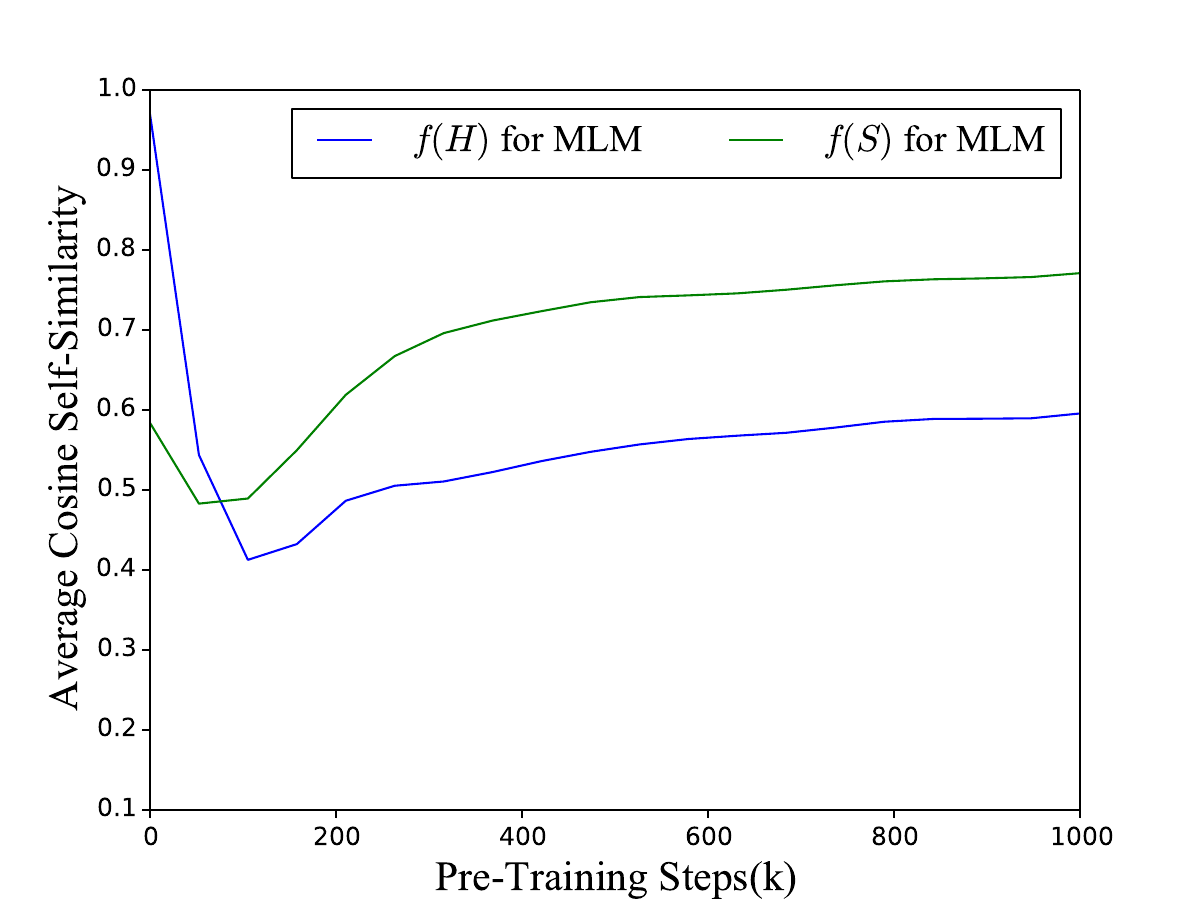}
    \caption{Average self-similarity of last hidden states and attention weights during MLM pre-training.}
    \label{fig:two}
\end{figure}Specifically, we sample 5,000 sentences from the validation set and evaluate the average self-similarity of last hidden states and attention weights under multiple checkpoints during the pre-training stage as shown in Figure~\ref{fig:two}.  We can notice that although \(f(S)\) and \(f(H)\) decrease at the beginning of pre-training, soon they start to rise gradually until the end of the training, and the similarities are always high throughout the training process. These fully demonstrates that in the MLM-based BERT, the overlap between different tokens and heads is strong and information confusion in model representations has become a problem worth to be solved.

\noindent \textbf{Training objectives.} To guide the model to produce more discriminative token representations and head representations, we propose a novel MTH pre-training objective, which combines original MLM with two novel Token Cosine Differentiation (TCD) objective and Head Cosine Differentiation (HCD) objective. 
Specifically, MTH applies the average cosine self-similarity of sampled last hidden states and attention weights as two auxiliary pre-training regularizers besides MLM \footnote{We have empirically verified that this below sampling strategy can greatly reduce the computational overhead with only a slight performance drop, comparing with regularizing all tokens and heads. In practice, we notice that setting \(n' = 50, m' = 2\) is fine on both BERT and DDRP, which will compress the additional computational overhead from about 30\(\%\) to 4\(\%\).}. For an input sequence \(S=[x_1,\ldots ,x_n]\), TCD samples \(n'\) \((n'<=n)\) tokens uniformly in sequence order to obtain a subsequence \(\tilde{S}=[\tilde{x}_1,\ldots,\tilde{x}_{n'}]\) and calculates the average cosine self-similarity of the subsequence's last hidden states as follows:

\begin{footnotesize}
\begin{gather}
\mathcal{L}_{T C D}=\frac{2}{n'(n'-1)} \sum_{i=1}^{n'-1} \sum_{j=i+1}^{n'} \cos \left(\tilde{h}_{i}, \tilde{h}_{j}\right), \tag{17}
\end{gather}\end{footnotesize}where \(\tilde{h}_i\) and \(\tilde{h}_j\) are the last hidden states of \(\tilde{x}_i\) and \(\tilde{x}_j\). For multiple heads \(H^l=[H_1^l,\ldots ,H_m^l]\) of a specific layer \(l\), HCD randomly samples \(m' \ (m'<=m)\) different heads \(\tilde{H^l}=[\tilde{H}_1^l,\ldots ,\tilde{H}_{m'}^l]\) (Note that HCD samples by layers, so sampled heads may be different across different layers.) and then calculate the average cosine self-similarity of attention weights of sampled heads as follows:

\begin{footnotesize}
\begin{gather}
\mathcal{L}_{HCD}=\frac{2}{Lm'(m'-1)} \sum_{l=1}^{L}\sum_{i=1}^{m'-1} \sum_{j=i+1}^{m'} \cos \left(\tilde{a}_{i}^l, \tilde{a}_{j}^l\right), \tag{18}
\end{gather}\end{footnotesize}\(\tilde{a}_{i}^l\) and \(\tilde{a}_{j}^l\) are the attention weights of the i-th head and j-th head in the sampled headset of the l-th layer. Ultimately, we define the global pre-training objective MTH as follows:

\begin{footnotesize}
\begin{gather}
\mathcal{L}_{\textbf{M}TH}=\mathcal{L}_{\textbf{M}LM} + \alpha_1\mathcal{L}_{\textbf{T}CD} + \alpha_2\mathcal{L}_{\textbf{H}CD}, \tag{19}
\end{gather}
\end{footnotesize}where \(\alpha_1\) and \(\alpha_2\) are hyperparameters.

\begin{table*}[!htbp]
\centering
\scalebox{0.9}{
\begin{tabular}{l|c|ccccccccc}
\hline
\textbf{Models} &\textbf{Steps}&\textbf{RTE}& \textbf{STS-B}& \textbf{MRPC}& \textbf{CoLA} &\textbf{SST-2}& \textbf{QNLI}& \textbf{QQP}& \textbf{MNLI}& \textbf{Avg.}\\
\hline
BERT (MLM) &1M&70.75&89.66&87.50&59.65&92.20&91.23&91.00&84.33&83.29\\
BERT-R (MLM)&1M&71.84&89.68&87.99&60.82&\textbf{92.66}&91.54&91.13&85.45&83.89\\
TUPE (MLM)&1M&68.59&89.61&86.02&\underline{62.82}&\textbf{92.66}&91.26&91.04&84.88&83.36\\
DEBERTA (MLM)&1M&\underline{73.28}&89.14&87.99&60.60&\textbf{92.66}&92.14&91.00&85.93&84.09\\
\hline
DDRP (MLM)&1M&72.20&\underline{90.01}&\underline{88.25}&\underline{62.82}&\underline{92.41}&\textbf{92.31}&\textbf{91.24}&\underline{86.02}&\underline{84.41}\\
DDRP (MTH)&1M&\textbf{75.09}&\textbf{90.41}&\textbf{88.72}&\textbf{63.36}&\textbf{92.66}&\underline{92.24}&\underline{91.22}&\textbf{86.22}&\textbf{85.00}\\
\hline
\end{tabular}
}
\caption{\label{tab:one}
Results on the development set of the GLUE benchmark for base-size pre-trained models. The best results are bolded, and the second results are underlined.}
\end{table*}

\begin{table*}[!htbp]
\centering
\scalebox{0.9}{
\begin{tabular}{l|c|ccccccccc}
\hline
\textbf{Approaches} &\textbf{Steps}&\textbf{RTE}& \textbf{STS-B}& \textbf{MRPC}& \textbf{CoLA} &\textbf{SST-2}& \textbf{QNLI}& \textbf{QQP}& \textbf{MNLI}& \textbf{Avg.}\\
\hline
BERT (MLM) &500k&68.23&88.92&86.74&57.05&91.97&90.41&90.74&83.41&82.18\\
BERT (MTH) &500k&\textbf{71.84}&\textbf{89.41}&\textbf{86.76}&\textbf{61.40}&\textbf{92.08}&\textbf{90.59}&\textbf{90.76}&\textbf{83.61}&\textbf{83.31} \\
\hline
BERT (MLM)&1M&70.75&89.66&87.50&59.65&92.20&\textbf{91.23}&91.00&84.33&83.29\\
BERT (MTH) &1M&\textbf{73.64}&\textbf{90.16}&\textbf{88.48}&\textbf{62.31}&\textbf{92.43}&91.21&\textbf{91.12}&\textbf{84.67}&\textbf{84.25}\\
\hline
DDRP (MLM)&1M&72.20&90.01&88.25&62.82&92.41&\textbf{92.31}&\textbf{91.24}&86.02&84.41\\
DDRP (MTH)&1M&\textbf{75.09}&\textbf{90.41}&\textbf{88.72}&\textbf{63.36}&\textbf{92.66}&92.24&91.22&\textbf{86.22}&\textbf{85.00}\\
\hline
\end{tabular}
}
\caption{\label{tab:two}
Development scores on GLUE benchmark. BERT (MLM/MTH) represents pre-trained BERT\(_{BASE}\) with MLM/MTH pre-training objective. DDRP (MLM/MTH) represents pre-trained DDRP\(_{BASE}\) with MLM/MTH pre-training objective.}
\end{table*}

\section{Experiments}

\subsection{Pre-training Text Corpora}

Follow \citet{devlin2018bert}, we use Wikipedia and BooksCorpus \citep{zhu2015aligning}, a roughly 16G uncompressed text corpus for pre-training.

\subsection{Baselines}

We compare DDRP with competitive pre-trained models. BERT \citep{devlin2018bert} equips Transformer \cite{vaswani2017attention} with parametric absolute position encoding. BERT-R uses the relative position encoding proposed by \citet{shaw2018self}, which couples relative distance information and directional information for modeling. TUPE \citep{ke2020rethinking} performs Query-Key dot product with different parameter projections for contextual information and positional information separately and then adds them up, plus the relative position biases like T5 \citep{raffel2019exploring}. 
DEBERTA \citep{he2020deberta} uses two vectors to encode content and position and uses disentangled matrices on their contents and relative positions respectively to compute the attention weights among words.

\subsection{Experimental Settings}
Following the previous practice, we use a base-size model for training, which consists of 12 Transformer encoder layers, each containing 12 heads with an input dimension of 768. During pre-training, we directly use the maximum training length of 512 without taking any form of random injection, and for examples less than 512 in length, we do not use the next document for padding. We remove Next Sentence Prediction (NSP) task and only keep Masked LM (MLM) as our pre-training task for all models unless noted otherwise. Considering that shorter documents may be missing semantics, we discard documents of length less than 8. We adopt the whole word masking strategy and split the whole words longer than 4 into individual subtokens. Following \citet{devlin2018bert}, we set the batch size to 256 sequences, the peak learning rate to 1e-4, and the training steps to 1M. We grid search \(\alpha_1\) and \(\alpha_2\) of TCD and HCD in \(\left\{ 0.01, 0.1, 1.0 \right\}\). Eventually, we set \(\alpha_1\) = 1.0 for TCD and set \(\alpha_2\) = 0.01 for HCD. All the models are implemented based on the code practice of BERT$\footnote{\url{https://github.com/google-research/bert}}$ in Tensorflow. We conduct all experiments on 16 Tesla-V100 GPUs (32G). All the pre-training hyperparameters are supplemented in Appendix~\ref{sec:appendix1}. To make a fair comparison, we implement BERT, BERT-R, TUPE, DEBERTA, and DDRP\footnote{Following \citet{shaw2018self} and \citet{raffel2019exploring}, we set \(r_s\) = 64 for all the relative position encoding models.} with the same pre-training hyperparameters and model configurations, which are consistent with vanilla BERT. 

\begin{figure*}[htbp]
	\centering
	\begin{minipage}{0.49\linewidth}
		\centering
		\includegraphics[width=8cm]{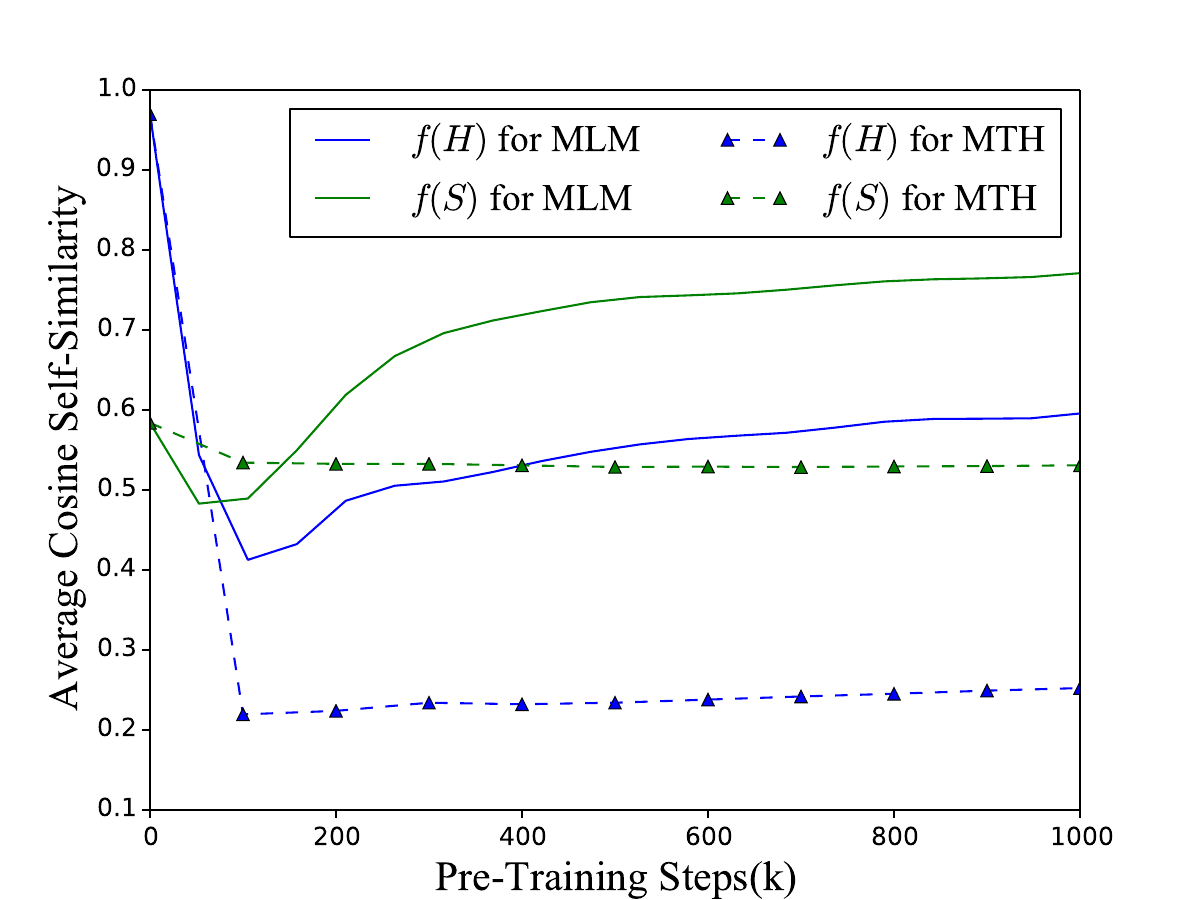}
	\end{minipage}
	\begin{minipage}{0.49\linewidth}
		\centering
		\includegraphics[width=8cm]{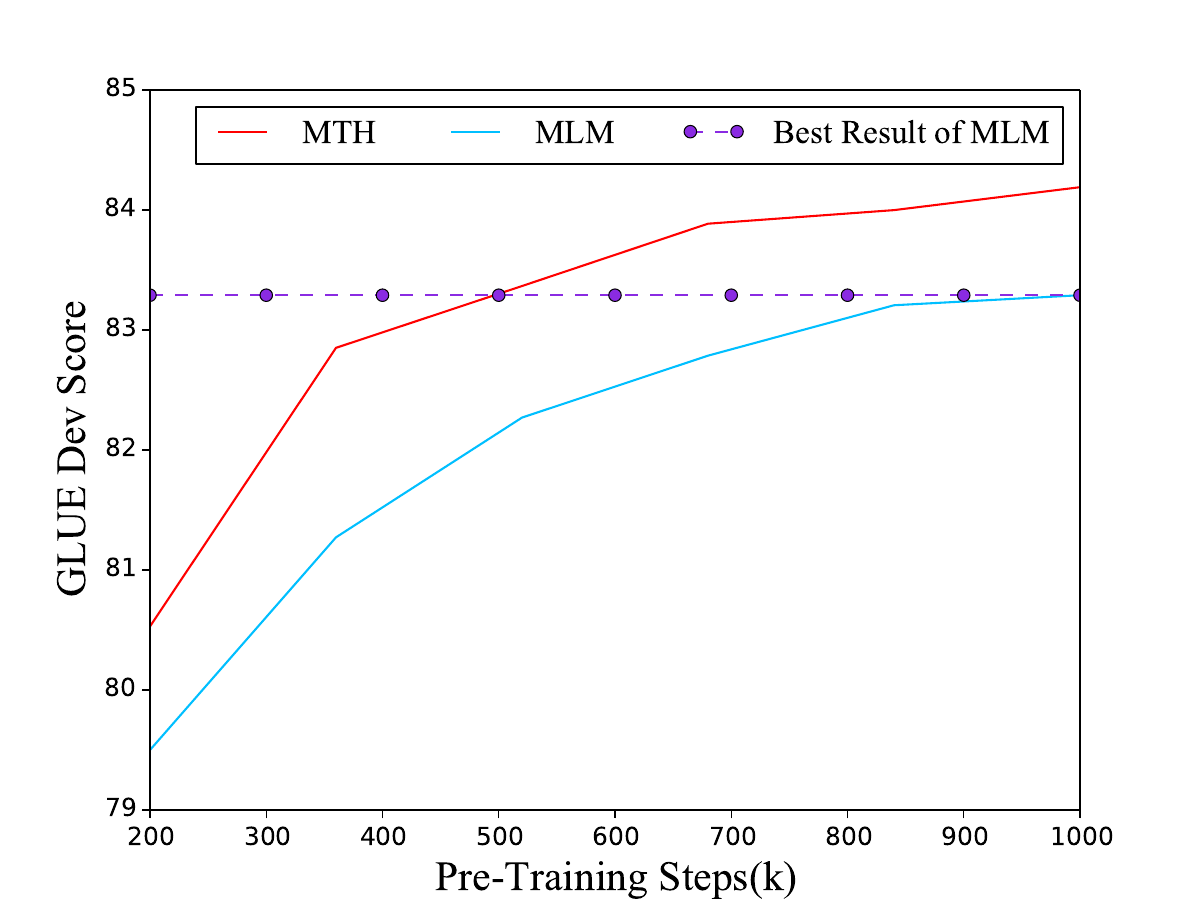}
	\end{minipage}
    \caption{The left figure (a) represents the trend of average cosine self-similarity of token representations and head representations during pre-training. The right figure (b) represents the trend of GLUE average score during pre-training.}
    \label{fig:four}
\end{figure*}

\subsection{Results on GLUE Benchmark}
We evaluate models on eight different English understanding tasks from General Language Understanding Evaluation (GLUE) benchmark \cite{wang2018glue}. The datasets cover four types of tasks: natural language inference (RTE, QNLI, MNLI), paraphrase detection (MRPC, QQP), linguistic acceptability (CoLA), and sentiment classification (SST-2).  For all experiments, STS-B and CoLA are reported by Pearson correlation coefficient and Matthews correlation coefficient, and other tasks are reported by Accuracy. All the fine-tuning hyperparameter configurations can be found in Appendix~\ref{sec:appendix2}. Following \citet{ke2020rethinking}, we fine-tune with five random seeds and report the median results.

\subsubsection{Comparing  Prior Competitive Models with DDRP}
The overall comparison results are shown in Table~\ref{tab:one}. 
Firstly, we can notice that all the various relative position encoding models  perform better than BERT, which proves that relative position encoding is a more competitive approach to encode position information.
Sencodly, it is easy to find that DDRP outperforms all the strong baselines, which demonstrates modeling relative position encoding by clarifying the originally confused relative distance and directional information more clearly is more effective.
Thirdly, DDRP pre-trained with MTH can consistently outperform BERT-R/DEBERTA by a 1.11/0.91 average GLUE score, which indicates that DDRP can be effectively compatible with better pre-training objectives to perform stronger.
Moreover, compared to BERT-R, DDRP introduces nothing in complexity while DEBERTA increases the computational cost about 25\(\%\), we also consider DDRP is a more time-efficient alternative than the recent state-of-the-art model DEBERTA (as analyzed in Sec.\ref{complexity}).

\subsubsection{Comparing MTH with MLM}
\label{Sec4.4.2}
As illustrated in Table~\ref{tab:two}, BERT (MTH) outperforms BERT (MLM) by a 0.96 average GLUE score and is consistently better on 7 out of 8 tasks. 
When combining MTH with strong DDRP, it still brings an improvement by 0.59 GLUE average score. 
Notably, BERT (MTH) can achieve better results compared with well-trained BERT (MLM) while only using 50\(\%\) training steps.  Since MTH utilizes cosine similarity and sampling strategy for the penalty, only a very slight computational cost is introduced. 
All the above statistics can fully verify that decreasing the similarity in model representations can effectively alleviate information overlap and increase representation diversity, which in turn leads to consistent and stable improvement across different model structures.

\section{Analysis and Discussion}
\subsection{Ablation Studies}
\textbf{Effect of DDRP.}
As shown in Table\ref{tab:one}, BERT-R 
outperforms BERT by 0.6 points on average. Based on BERT-R, our proposed DDRP outperforms BERT-R by 0.52 points averagely without imposing additional computational costs. It is worth noting that compared to BERT-R, DDRP helps a great deal on low-resource tasks, such as CoLA, while further improving the performance on high-resource tasks, such as QNLI and MNLI. These fully demonstrate utilizing separate parametric vectors to represent distances and directions, two apparently heterogeneous information, can be beneficial to the model, and further justifies the value of dissociating confusing information that is confounded in similar spaces.

 \noindent\textbf{Effect of TCD and HCD.}
 MTH brings in two additional TCD and HCD regularizers besides the original MLM task. To further evaluate the relative contributions of the HCD and TCD, we develop one variation, which is BERT pre-trained with MLM and TCD. Table~\ref{tab:six} summarizes the results on the base-size models. Firstly, it shows a 0.42 average GLUE score drop when HCD is removed from MTH, especially on MRPC, CoLA, and MNLI. Secondly, there is a 0.54 average GLUE score drop when TCD is progressively removed, especially on RTE, STS-B, and CoLA. These results indicate that both TCD and HCD regularizers play a crucial role in improving performance.
 
 \begin{table*}[!htbp]
\centering
\scalebox{0.9}{
\begin{tabular}{l|ccccccccc}
\hline
\textbf{Model} &\textbf{RTE}&\textbf{STS-B}&\textbf{MRPC}&\textbf{CoLA}&\textbf{SST-2}&\textbf{QNLI}&\textbf{QQP}&\textbf{MNLI}& \textbf{Avg.}\\
\hline
MTH&\textbf{73.64}&90.16&\textbf{88.48}&\textbf{62.31}&\textbf{92.43}&91.21&\textbf{91.12}&\textbf{84.67}&\textbf{84.25}\\
\ w/o HCD&73.28&\textbf{90.41}&87.25&60.85&92.23&\textbf{91.37}&91.00&84.22&83.83\\
\ w/o TCD\&HCD &70.75&89.66&87.50&59.65&92.20&91.23&91.00&84.33&83.29 \\
\hline
\end{tabular}
}
\caption{\label{tab:six}
Ablation study for MTH. Note that MTH (w/o TCD\&HCD) equals simply using MLM in pre-training.}
\end{table*}
 
\subsection{Analysis on MTH}
\label{Sec.5.2}
To further understand why MTH works, we compare MLM and MTH in terms of average self-similarity of token representations and head representations and performance during pre-training in Figure~\ref{fig:four}. As shown in Figure~\ref{fig:four}.(a), it is easy to find that MTH's average self-similarity is much lower than MLMs'. We can also clearly notice from  Figure~\ref{fig:four}.(b) that the average GLUE score of MTH is always about one point higher than MLM's during the whole pre-training process. These confirm that (i) the differentiation of tokens and heads is important for model optimization; (ii) MTH can help to produce more discriminative token representations and head representations, extend the representation space of tokens and heads, and thus improve the performance.

\subsection{Analysis on DDRP}
\label{ddrp}
\begin{figure}[htbp]
    \centering
    \includegraphics[width=8cm]{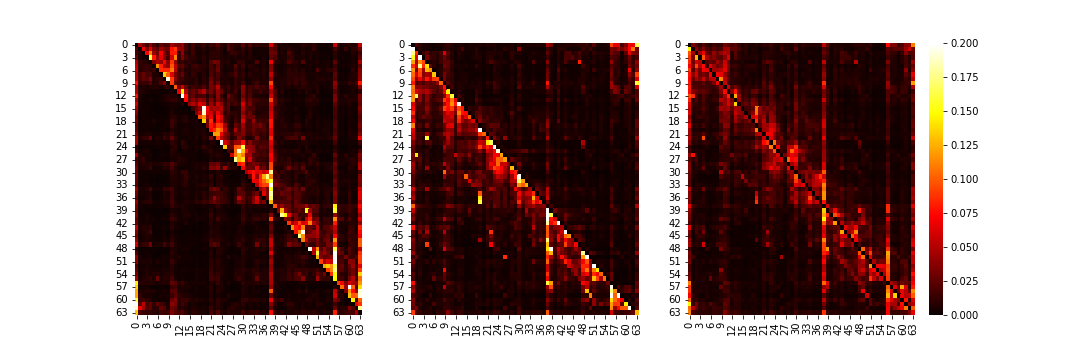}
    \caption{Attention visualization for a sampled batch of sentences. From left to right is the attention visualization for group 1, group 2, and global, respectively.}
    \label{fig:three}
\end{figure}

In this subsection, we intend to analyze the attention maps of DDRP as a way to investigate why making only slight modifications on BERT-R can bring great gains. To better examine and explain the ability of DDRP to capture information in both left and right directions, we divide multiple heads into two groups evenly, where group 1 consists of the heads that focus most on the right side, and group 2 consists of the heads that focus most on the left side. As shown in Figure~\ref{fig:three}, it is easy to observe a distinct upper triangle effect in group 1 and a distinct lower triangle effect in group 2, which indicates that DDRP may allow the model to be more precise in the perception of direction, a piece of information that is crucial to understanding semantics.
To further confirm our point, we sample 5,000 sentences from the validation set and count the percentage of sentences with upper and lower triangular effects according to Algorithm 1 (more details can be seen in Appendix~\ref{sec:appendix3}). 
It is observed that 92.11\(\%\) of sentences have an up-down triangle effect.
We also count the percentage for BERT-R with the same process and observe only 78.94\(\%\) of the sentences have an up-down triangle effect. 
All the phenomena and statistics fully reveal that DDRP 
can make different heads focus on the token information interaction in different directions and reduce confusion between heads, thus improving the effectiveness and rationality of the model.

\subsection{Complexity Analyses}
\label{complexity}
\textbf{DDRP.} Compared with BERT, DDRP introduces additional parameters: \(D \in R^{3\times d}\) and \(K^{rd}\in R^{r_s\times d}\). The total increase in parameters is \(3 \times d +   r_s \times d\). For base-size model \((D=768,L=12,S=512,N=12,d=64)\)\footnote{N is the number of head.}, the total increase amounts to 0.0043M, which is negligible.  
Compared with BERT, the additional computational complexity for both BERT-R and DDRP is \(O(SD)
\). 
Since DEBERTA equips different heads with unshared \(K^r\)s, the additional computational complexity for DEBERTA is  \(O(NSD)\). 
Overall, BERT-R and DDRP increase the computational cost about 5\(\%\), and DEBERTA increases the computational cost about 30\(\%\). 
Although DDRP introduces a slight computational cost compared to BERT, it is more time-efficient than DEBERTA and outperforms all the above models.

\noindent\textbf{MTH.} Since the two regularizers of TCD and HCD are based on cosine similarity and sampling strategy, they do not introduce too much computational cost. Compared with MLM, MTH only increases a slight computational cost of about 4\(\%\) while bringing excellent improvement.

\nocite{devlin2018bert,liu2019roberta,lan2019albert,yang2019xlnet,clark2020electra,vaswani2017attention,shaw2018self,raffel2019exploring,ke2020rethinking,kovaleva2019revealing,he2020deberta,press2020shortformer,zhu2015aligning,cui2020revisiting,,loshchilov2018fixing,sun2019ernie,wang2018glue,joshi2020spanbert,xu2020clue,jiang2020convbert,dai2019transformer,mu2017all,mimno-thompson-2017-strange,mikolov2013efficient,pennington2014glove,michel2019sixteen,shen2018disan,vu2016bi,fuller2002arrow}

\section{Conclusion}
In this work, we analyze and identify potential information confusion in the relative position encoding and model representations, respectively, and design two novel techniques to address these problems: DDRP (Decoupled Directional Relative Position) encoding and MTH (MLM with TCD and HCD) pre-training objectives.
Specifically, DDRP decouples relative distance features and directional features to eliminate unnecessary randomness in the self-attention module. 
MTH utilize TCD and HCD as two regularizers to supervise the model to always maintain a certain level of critical thinking. 
The experimental results show that DDRP achieves better performance compared with various relative position encoding models and MTH outperforms MLM by a large margin. 
We believe that reducing information confusion in representation learning may have broader application scenarios, and leave this area of exploration
for future work.

\section{Limitations}
Our limitations lie in inducing additional
computational costs. 
Compared with BERT, the additional computational complexity for DDRP is \(O(SD)
\)\footnote{Here, \(S\) is the input sequence length and \(D\) is the dimension of token representations.}, which is reflected in the 5\(\%\) increase in computational cost.
Compared with MLM, MTH with sampling strategy increases the computational cost by about 4\(\%\).

\bibliography{anthology,eacl2023}
\bibliographystyle{acl_natbib}

\clearpage

\appendix

\section{Appendix A. Hyperparameters for Pre-Training}
\label{sec:appendix1}

As shown in Table~\ref{tab:ten}, we list the pre-training hyperparameter configurations. To make a fair comparison, all models' pre-training hyperparameter configurations in our experiments are identical to vanilla BERT \cite{devlin2018bert}.

\begin{table}[!h]
\centering
\begin{tabular}{lr}
\hline
\textbf{Hyperparameter} &\\
\hline
Vocab size &3,0522 \\
Hidden size &768\\
Attention heads &12\\
Layers&12 \\
Training steps& 1M \\
Warmup ratio&0.01 \\
Batch size&256\\
Learining rate&1e-4 \\ 
Adam $\epsilon$ &1e-6\\
Adam $(\beta_1,\beta_2)$&(0.9,0.999) \\
Learning rate schedule&linear \\
Weight decay&0.01 \\
Clip norm &1.0\\
Dropout &0.1\\
\hline
\end{tabular}
\caption{Hyperparameters used for pre-trained models.}
\label{tab:ten}
\end{table}

\section{Appendix B. Hyperparameters for Fine-Tuning}
\label{sec:appendix2}

As shown in Table~\ref{tab:seven}, we enumerate the hyperparameter configurations to fine-tune the tasks on the GLUE benchmark \citep{wang2018glue}. We grid search these fine-tuning hyperparameter configurations for all models. Following the BERT, we do not show results on the WNLI GLUE task for the Dev set results.

\begin{table}[!h]
\centering
\scalebox{0.9}{
\begin{tabular}{lr}
\hline
\textbf{Hyperparameter} & \textbf{GLUE}\\
\hline
Batch size&$\left\{ 16, 32 \right\}$ \\
Learining rate&$\left\{ 1e-5, 2e-5, 3e-5 \right\}$ \\ 
Epoch& \(\left\{4 , 6 \right\}\)  \\
Adam $\epsilon$ &1e-6\\
Adam $(\beta_1,\beta_2)$&(0.9,0.999) \\
Learning rate schedule&linear \\
Weight decay&0.01\\
Clip norm &1.0\\
Dropout &0.1\\
Warmup ratio & 0.1\\
\hline
\end{tabular}}
\caption{Hyperparameters used for fine-tuning on the GLUE benchmark.}
\label{tab:seven}
\end{table}

\section{Appendix C. Details for Up-Down Triangle Effect }
\label{sec:appendix3}
Here we provide more details for the up-down triangle effect (in Sec.~\ref{ddrp}). It is rather difficult and non-intuitive to analyze the directional information in $12$ different attention heads.
Since previous studies have considered to group multiple heads in the self-attention module \cite{2018psm,2021paybetter}, we thereby attempt to divide the heads into two groups evenly. 
Specifically, we divide the heads that focus more on the right side information into group 1 and the heads that focus more on the left side information into group 2, wishing to reveal the directional information encoded in attention weights more explicitly. We experimentally verified that this grouping approach could provide a better presentation of the directional information. Therefore, we combined this approach to conduct a comparative analysis of DDRP and BERT-R to demonstrate the more powerful directional perception of DDRP.

As illustrated in Figure~\ref{fig:three}, group 1 is more focused on the right information (greater attention values in the upper triangle region) and group 2 is more focused on the left information (greater attention values in the lower triangle region). To further analyze the universality of this phenomenon, we design the Algorithm~\ref{alg:algorithm1} to quantitatively analyze the ability of DDRP to capture information on the both left and right sides. To make a fair comparison, we also conduct the same process for BERT-R.

\begin{algorithm}[!htpb]
\centering
\begin{footnotesize}
\caption{Count up-down triangle percentage} 
{\bf Require:} \(N\) : total number of sentences; \(n\) : total number of sentences that have been processed; \(ms\) : maximum sentence length; \(mn\) : total number of sentences that match the upper and lower triangle; \(t\) : the threshold value that satisfies the up-down triangular effects; \(amp\) :  attention map obtained by averaging attention maps in specific group.\\
\begin{algorithmic}[1]
\State \textbf{Initialize} \(ms \leftarrow 64\), \(n \leftarrow 1\), \(mn \leftarrow 0\), \(t \leftarrow 0.7\) 
\While{\(n\le N\)} 
\State Divide heads in equal with greater attention
values in the (upper/lower) triangle region into (group 1/group 2) and obtain (\(amp1/amp2\)).
  		\State // prepare for the upper and lower triangle
  		\State Sum the values in upper and lower triangles of \(amp1\) respectively and obtain \(amp1_{up}\), \(amp1_{down}\).
        \State Sum the values in upper and lower triangles of \(amp2\) respectively and obtain \(amp2_{up}\), \(amp2_{down}\).
        \State  \(amp1_{up}\leftarrow amp1_{up}/ms\), \(amp1_{down}\leftarrow amp1_{down}/ms\)
        \State \(amp2_{up}\leftarrow amp2_{up}/ms\), \(amp2_{down}\leftarrow amp2_{down}/ms\)
        \State // compute for the upper and lower triangle
        \If{\(amp1_{up} \ge t\) and \(amp2_{down} \ge t\)}
\State \(mn \gets mn+1\) 
\Else
\State \textbf{Continue}
\EndIf
  		\State \(n \gets n+1\)
\EndWhile
\State \Return \(float(mn/N)\)
\end{algorithmic}
\label{alg:algorithm1}
\end{footnotesize}
\end{algorithm}

\end{document}